\documentclass[10pt,twocolumn,letterpaper]{article}

\usepackage{wacv}              

\usepackage{graphicx}
\usepackage{amsmath}
\usepackage{amssymb}
\usepackage{booktabs}
\usepackage{times}     

\usepackage{multirow}  

\usepackage{tabu}                      
\usepackage{mathptmx}                  
\usepackage{amssymb}
\usepackage{pifont}
\usepackage{bm}
\usepackage{balance}
\usepackage{tikz}
\usepackage{threeparttable}
\usepackage[pagebackref,breaklinks,colorlinks]{hyperref}

\usepackage[capitalize]{cleveref}
\crefname{section}{Sec.}{Secs.}
\Crefname{section}{Section}{Sections}
\Crefname{table}{Table}{Tables}
\crefname{table}{Tab.}{Tabs.}

\def\wacvPaperID{2735} 
\def\confName{WACV}
\def\confYear{2025}

\begin{document}

\definecolor{ao(english)}{rgb}{0.0, 0.5, 0.0}

\newcommand{\cmark}{\ding{51}}%
\newcommand{\xmark}{\ding{55}}%

\newcommand{\NYU}{{NYU Depth V2}\xspace}
\newcommand{\ArkitScenes}{{ARKitScenes}\xspace}
\newcommand{\ddff}{{DDFF12}\xspace}
\newcommand{\mobiledepth}{{Mobile Depth}\xspace}

\newcommand{\pipeline}{\textsc{HybridDepth}\xspace}
\newcommand{\para}[1]{\noindent  {\bf #1}}

\newcommand{\ashkan}[1]{{\color{cyan}\textbf{[Ashkan: \textit{#1}]}}}
\newcommand{\tian}[1]{{\color{orange}\textbf{[Tian: \textit{#1}]}}}
\newcommand{\HS}[1]{{\color{blue}\textbf{[HS: \textit{#1}]}}}

\newcommand{\circlednumber}[1]{%
  \tikz[baseline=(char.base)]{
    \node[shape=circle, draw, inner sep=1pt, minimum size=1em] (char) {\textbf{#1}};}}

\title{HybridDepth: Robust Metric Depth Fusion by \\ Leveraging Depth from Focus and Single-Image Priors}

\author{Ashkan Ganj\\
Worcester Polytechnic Institute\\
{\tt\small aganj@wpi.edu}
\and
Hang Su\\
Nvidia Research\\
{\tt\small hangsu@nvidia.com}
\and
Tian Guo\\
Worcester Polytechnic Institute\\
{\tt\small tian@wpi.edu
}
}

\maketitle
\begin{abstract}
We propose \pipeline, a robust depth estimation pipeline that addresses key challenges in depth estimation, including scale ambiguity, hardware heterogeneity, and generalizability. 
\pipeline leverages focal stack, data conveniently accessible in common mobile devices, to produce accurate metric depth maps. 
By incorporating depth priors afforded by recent advances in single-image depth estimation, our model achieves a higher level of structural detail compared to existing methods. We test our pipeline as an end-to-end system, with a newly developed mobile client to capture focal stacks, which are then sent to a GPU-powered server for depth estimation.

Comprehensive quantitative and qualitative analyses demonstrate that \pipeline outperforms state-of-the-art (SOTA) models on common datasets such as \ddff and \NYU. 
\pipeline also shows strong zero-shot generalization. When trained on \NYU, \pipeline surpasses SOTA models in zero-shot performance on \ArkitScenes and delivers more structurally accurate depth maps on \mobiledepth. The  code is available at \href{https://github.com/cake-lab/HybridDepth/}{https://github.com/cake-lab/HybridDepth/}.
\end{abstract}

\vspace{-5mm}

\section{Introduction}
\label{intro}

Depth estimation is a critical task in computer vision, with applications~\cite{mahjourian2024multimodalobjectdetectionusing} spanning autonomous driving, augmented reality~\cite{10.1007/978-3-031-60881-0_19, 10.1007/978-3-031-35822-7_33}, and robotics~\cite{zhang2024deeppointmap}. 
While hardware like LiDAR and Time-of-Flight (ToF) sensors are commonly used for more expensive applications, they are often not available for mobile and consumer-specific applications. 
This has led to extensive research on vision-based depth estimation methods that rely solely on cameras. Monocular depth estimation models have gained popularity due to their simplicity and minimal hardware requirements; however, these models frequently suffer from scale ambiguity, where depth estimations vary across different zoom levels. 
Additionally, single-image models, such as ZoeDepth~\cite{https://doi.org/10.48550/arxiv.2302.12288} and ECoDepth~\cite{patni2024ecodepth}, struggle to generalize well to real-world conditions, as demonstrated by recent work~\cite{10.1145/3638550.3641122} using \ArkitScenes~\cite{dehghan2021arkitscenes}.

Depth-from-focus (DFF) methods, which leverage focal stack information, provide a promising alternative by offering more robust depth cues and generating reliable metric depth estimates using only cameras. However, existing DFF-based models like DFV~\cite{3eb3b6e31bc743c68f1e7439bd4f4799} tend to produce noisy estimations in texture-less or challenging regions and can't generalize well, limiting their overall effectiveness. On the other hand, relative depth models, while capable of generalizing well to unseen data and capturing fine details in depth maps, do not provide metric information, making them unsuitable for applications that require metric depth values. 

In this work, we address the above mentioned limitations by combining DFF with relative depth models. While relative depth models excel in generalization and structural accuracy, DFF provides reliable metric depth information without requiring specialized hardware. By integrating these two approaches, we have the potential to leverage the strengths of both, delivering robust and accurate depth estimation that is both scalable and applicable to a wide range of real-world scenarios.
\begin{figure}[t]
    \centering
        \includegraphics[width=\columnwidth]{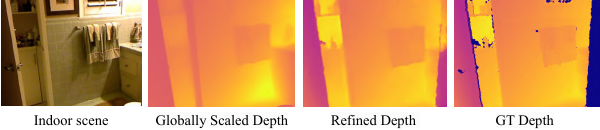}
\caption{\pipeline produces globally scaled depth maps, and refines them to further correct errors and enhance details.
}

\label{fig:pipelineSteps}
\vspace{-10pt}
\end{figure}

Intuitively simple, we have to address the challenge of effectively combining focal stack information with relative depth models to maintain the \emph{structural accuracy and generalization capabilities} of relative depth estimation, while incorporating the \emph{reliable metric depth information} from DFF. 
Toward this end, we propose an end-to-end solution, called \pipeline, that fuses DFF and single-image priors to achieve robust metric depth estimation.

\pipeline is designed to deliver zero-shot performance, \ie to effectively generalize to unseen data or scenes,  
by employing a three-stage approach. First, we capture the outputs from both the relative depth and metric depth models. Next, we perform a global scaling that aligns the relative depth output with the absolute scales provided by the metric information from DFF. Finally, we apply a deep learning-based refinement layer to fine-tune the intermediate depth map (i.e., globally scaled depth map), smoothing out any inconsistencies and enhancing the overall accuracy of the depth estimations.
Figure~\ref{fig:pipelineSteps} provides an example of how \pipeline refines the globally scaled depth map.

We conduct comprehensive experiments to evaluate our method on two real-world focal stack datasets, \ddff and \mobiledepth\footnote{Only qualitative results are shown due to lack of ground truth depth.}.
Due to the lack of focal stack datasets, we evaluate \pipeline on additional datasets, including \NYU and \ArkitScenes, with synthesized focal stacks. 
Our results demonstrate that \pipeline establishes new SOTA on depth estimation with focal stacks, as well as improved generalization capabilities. 
The qualitative zero-shot results on \mobiledepth and \ArkitScenes show that \pipeline also generalizes well.  

For example, \pipeline achieves a 10.5\% and 6.1\% improvement of MSE and RMSE on \ddff, and 6.5\% and 7.1\% improvement of RMSE and AbsRel on the \NYU dataset, compared to DFV~\cite{3eb3b6e31bc743c68f1e7439bd4f4799}. 
Moreover, our zero-shot performance on the \ArkitScenes dataset shows a 43\% improvement compared to SOTA methods~\cite{depthanything}.
We also conduct comparisons with single-image depth estimation approaches such as Depth Anything~\cite{depthanything} to demonstrate the significant advantage of utilizing the focal stack information on mobile devices. 

In summary, our main contributions are as follows.  
\begin{itemize}
    \setlength\itemsep{-0.4em}
    \item We design and implement an end-to-end pipeline \pipeline that demonstrates the feasibility and benefits of fusing focal stack information with single-image priors in achieving robust metric depth estimation. 
    \emph{Relevant research artifacts will be open-sourced.}
    \item \pipeline establishes new SOTA on DFF-based depth estimation, with a compact model of 240 MB and an inference time of 20 ms, which is 19.2\% of the Depth Anything size and 2.85X faster. 
    \item \pipeline showcases strong generalization performance on \mobiledepth and the AR-specific dataset \ArkitScenes.
    \item We demonstrate the significant advantage of \pipeline over single-image depth estimation approaches, offering a viable and attractive alternative for mobile applications. 
\end{itemize}

\begin{figure*}[ht]
    \centering
        \includegraphics[width=\linewidth]{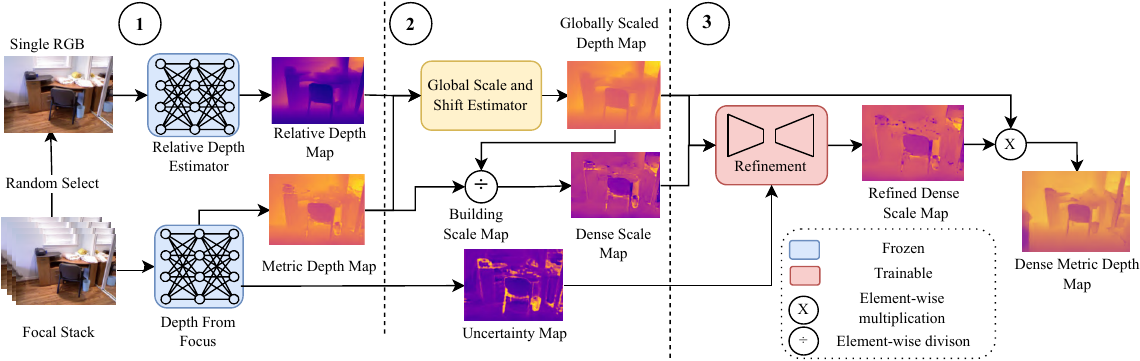}
    \caption{An overview of \pipeline which consists of three stages: (1) capture a focal stack and pass the frames through two branches; (2) calculate scale and shift based on estimated relative and metric depth maps using least-squares fitting; (3) input a globally scaled depth map and a processed version of the Metric DFF branch output to the refinement model to output the updated scale map, which will be applied to the globally scaled depth map to get the final depth map.
    }
\label{fig:model_arch}
\end{figure*}

\section{Related Work}
\label{sec:background}

\textbf{Single-image depth estimation} outputs depth maps given only single input images. Due to the inherent ill-posedness of the task, such approaches usually adopt an affine-invariant depth map formulation. 
This specific formulation enables training on large datasets, and has been proven to be key to model generalization. 
For instance, models like Midas~\cite{birkl2023midas,Ranftl2020}, DPT~\cite{Ranftl2021}, and Depth Anything~\cite{depthanything} have made a good progress in zero-shot performance by using a novel scale-invariant loss function, enabling training on datasets captured with various hardware devices, and new way of using data from different domains. Also, these models are very good at maintaining structural and segmentation accuracy.
Our work leverages the recent advancements in relative depth in zero-shot performance~\cite{depthanything} as the basis for achieving robust metric depth performance.

\textbf{Metric depth estimation} aims to provide exact depth values in physical units, such as meters. 
While this was traditionally only feasible with stereo or video approaches, models like Depth Anything~\cite{depthanything}, ZoeDepth~\cite{https://doi.org/10.48550/arxiv.2302.12288}, AdaBin~\cite{9578024}, and LocalBin~\cite{10.1007/978-3-031-19769-7_28} have shown success with single image inputs. 
However, such approaches usually degrade significantly for unseen data~\cite{10.1145/3638550.3641122} and unknown camera models~\cite{yin2023metric}. 
In contrast, our work relies less on visual data and, therefore, can circumvent the current problems of metric depth estimation.

\textbf{Depth from focus (DFF)} estimates depth by identifying the focus distance at which each pixel is most sharply defined, while areas outside the focal plane appear blurry, creating a circle of confusion (CoC). Deep learning-based methods~\cite{Defocus_2020_CVPR, 3eb3b6e31bc743c68f1e7439bd4f4799, hazirbas18ddff, Wang-ICCV-2021} use convolutional neural networks (CNNs) and neural networks to find the best focal plane for each pixel from a limited number of images. However, DFF can be noisy when suitable focal planes are missing from the data or when dealing with texture-less regions. Our design effectively mitigates these limitations by combining DFF with relative depth information and refining the depth map, ensuring accurate and stable estimations.


\section{\pipeline}
\label{sec:methodology}
We propose \pipeline, a three-stage pipeline that achieves generalizable metric depth without subjecting to the scale ambiguity issue. 
The design of \pipeline is inspired by ViDepth~\cite{wofk2023videpth}, which uses IMU sensors. 
Unlike ViDepth, \pipeline relies only on RGB inputs, fulfilling the goal of providing highly detailed metric depth maps without the burden of additional sensors on mobile applications. 
Our key idea is to \emph{leverage a well-generalized relative depth model as the basis and then convert the relative depth to metric depth with the help of a DFF model}.
The focal stack provides precise depth information using just a camera~\cite{3eb3b6e31bc743c68f1e7439bd4f4799}, and relative depth estimation is known for its ability to generalize well across different scenes while maintaining strong structural accuracy~\cite{depthanything, Ranftl2020}. By combining these two methods, we show that \pipeline can benefit from both worlds and outperform both methods~\cite{Ranftl2020, 3eb3b6e31bc743c68f1e7439bd4f4799}. 

Figure~\ref{fig:model_arch} describes the overall architecture of \pipeline. 
In essence, the pipeline performs pixel-wise linear transformation to convert each pixel's relative depth to metric depth.
The pipeline begins by processing the input focal stack. 
A randomly selected frame within the focal stack is passed through the relative depth estimator to generate a relative depth map, while the entire focal stack is fed into the DFF branch to produce a metric depth map~\circlednumber{1}. 
These outputs are then aligned using a global scale and shift estimation process~\circlednumber{2}. 
Finally, we address the remaining scale errors that occurred in \circlednumber{2} with a trainable scale refinement network~\circlednumber{3}. 
In the following sections, we provide details about each of the three modules within our pipeline.

\subsection{Capturing Relative and Metric Depth \texorpdfstring{\protect\circlednumber{1}}{}}

The first step of \pipeline consists of two key modules, a \emph{single-image relative depth estimator} and a \emph{DFF metric depth estimator}, to generate the intermediate data for the later metric conversion and refinement. Note that even though the DFF module can output the metric depth map, the depth map quality can be low and lack details, as we will show later in the evaluation.  

\para{Relative Depth Estimator.} This module generates a dense relative depth map, serving as the foundation for our depth estimation process. It is designed to utilize a pre-trained model that takes a single RGB image as an input and produces a dense relative depth map. The reason we use a pre-trained model like  Depth Anything\footnote{Note that Depth Anything can also output metric depth directly, which we outperform as will be shown in the evaluation section.}~\cite{depthanything} is because it is trained on large amounts of datasets and, therefore, has a potential for good zero-shot performance. 

\para{DFF Metric Depth Estimator.}  This module provides the crucial metric information needed to convert the relative depth map into a generalizable metric depth map. Unlike other methods that rely on external sensors~\cite{10.1145/3517260}, our approach is entirely visual-based. By leveraging focus cues from a \emph{focal stack}, consisting of images captured at different focus distances, this model assigns each pixel to its corresponding focus distance, yielding the metric depth for that specific pixel. For this module, we employ DFV\footnote{Directly using models like DFV will not give us the generality as we will show in the evaluation.}~\cite{3eb3b6e31bc743c68f1e7439bd4f4799}, a lightweight SOTA model that produces accurate metric depth for the DDFF12 dataset.

\subsection{Relative and Metric Depth Fusion \texorpdfstring{\protect\circlednumber{2}}{}}
\label{subsec:Global_alignment}

\para{The Global Scale and Shift Alignment.} This module performs the linear transformation of the relative depth map to a globally scaled depth map using Equation~\eqref{eq:scale_shift_eq}. 
\begin{equation}
    \label{eq:scale_shift_eq}
    \text{Metric Depth} = \text{Scale} \times \text{Relative Depth} + \text{Shift}
\end{equation}
Here, the \emph{Scale} and \emph{Shift} parameters are obtained via the least-square fitting~\cite{Ranftl2020}  using the metric and relative depth maps obtained from \circlednumber{1}. Applying the scale value globally for all pixels brings relative depth values to a correct order of magnitude, while applying global shift can help undo potential bias or offset in the original estimation.
Unlike prior work~\cite{https://doi.org/10.48550/arxiv.2302.12288, depthanything} that often alter the fundamental depth relationships, scaling relative depth directly allows us to maintain such relationships, and therefore avoid distortions and ensure high-quality depth maps.

\para{Building Dense Scale Map.}  
To construct the dense scale map, we compute the pixel-wise scale difference between the globally scaled depth map (produced in  \circlednumber{2}) and the DFF branch's depth output (from \circlednumber{1}). This process involves a straightforward pixel-level division, where each pixel in the globally scaled depth map is divided by the corresponding pixel in the DFF depth map. 
The resulting scale map captures the local scale variations, which will later be used in the refinement layer \circlednumber{3} to generate a refined dense scale map. 
This pixel-wise division ensures that the refinement layer can adjust each pixel's depth value according to its specific scale discrepancy.

\subsection{Scale Refinement Layer \texorpdfstring{\protect\circlednumber{3}}{}}
\label{subsec:refinement}
Global scale and shift alignment can introduce errors, as it attempts to convert the entire relative depth map into metric depth using only two parameters. This oversimplification may result in inaccuracies across specific pixels and regions. Our experiments suggest that certain areas of the globally scaled depth maps can benefit from localized scale refinements. To address this, we introduce a refinement layer that applies pixel-wise scale corrections to the globally scaled depth map, utilizing the scale map and uncertainty map derived from the DFF module.

Our refinement process leverages a customized version of MiDaS-small~\cite{Ranftl2022} to correct pixel-wise scale errors. Specifically, this refinement model leverages the globally scaled depth map, the DFF-derived scale map, and the uncertainty map and outputs a refined dense scale map, which consists of pixel-wise depth scale adjustments. The uncertainty map allows the refinement layer to account for uncertainties from DFF, effectively guiding the model in weighting the influence of each pixel in scale refinement. This addition of uncertainty awareness empowers the refinement model to make more informed, precise adjustments, significantly improving the final depth map’s accuracy and robustness. By integrating visual cues, pixel-wise scale refinement, and uncertainty-driven adjustments, this layer offers a novel and highly effective approach to refining depth maps, delivering enhanced metric precision without reliance on external sensors.

\section{Implementation Details}
Our model is implemented using the PyTorch framework. For real-world testing, we developed a mobile client using the Android Camera2 API to capture focal stacks, coupled with an edge server equipped with an NVIDIA RTX 4090 GPU for inference.

\subsection{Training}
\label{subsec:training}

We use Depth Anything pre-trained weights in all experiments to leverage its generalization. For the DFV module, we either use pre-trained weights (\ddff, \mobiledepth, and \ArkitScenes) or train it from scratch (\NYU). For the zero-shot experiments on \ArkitScenes, we use both the pre-trained DFV models from \NYU and \ddff, with the refinement layer trained on \NYU. For \mobiledepth, we use the pre-trained DFV and train the refinement layer on \NYU.

All models are trained using the AdamW optimizer, with dataset-specific hyperparameters and batch sizes. Further training details, including hyperparameters, input sizes, and augmentations for each dataset, can be found in Supplementary Material.

\subsubsection{Loss Function}
\label{subsec:loss_func}

Prior work like ViDepth~\cite{wofk2023videpth}  utilizes L1 loss as their regression task loss function. However, we know that L1 loss is sensitive to changes in distance ranges, hindering performance on unseen data~\cite{eigen2014depth}. To
address this issue, we adopt the \emph{scale-invariant} loss function $L_{\text{SILog}}$ proposed in \cite{eigen2014depth}. 
Additionally, we integrate a multi-scale gradient loss function $L_{\text{grad}}$ to enhance visual quality and sharpness while preserving image boundaries as much as possible. Our overall loss function $L$ is as follows:

\begin{equation}
    L = L_{\text{SILog}} + 0.5 \times L_{\text{grad}},
\end{equation}

where \(L_{\text{SILog}}\) is defined as:

\begin{equation}
    L_{\text{SILog}} = 10 \times \sqrt{\text{var}(g) + \beta \times (\text{mean}(g))^2}.
\end{equation}

Here, \( g = \log(d + \alpha) - \log(d_{gt} + \alpha) \), with \(\alpha\) being a small constant to prevent undefined logarithmic operations, and \(\beta\) serving as a scaling factor for the mean squared term. $\alpha$ and $\beta$ are set to $1e^{-7}$ and $0.15$. 


\(L_{\text{grad}}\) is defined as: 

\begin{equation}
    L_{\text{grad}} = \frac{1}{HW} \sum_{s=1}^{4} \sum_{i,j} \left| \nabla_s d_{i,j} - \nabla_s d_{gt,i,j} \right|,
\end{equation}

where \(\nabla\) denotes the first-order spatial gradient operator, \(s\) indicates the scale factor for multi-scale analysis, \(d\) represents the predicted depth map, and \(d_{gt}\) is the ground truth depth map. \(H\) and \(W\) are the height and width of the depth map, respectively. This composite loss function aims to optimize both the scale-invariant and gradient-based aspects of the predicted depth map, enhancing accuracy and geometrical information.

\subsection{Data Synthesizing}
\label{sec:data_synthesize}

The ability to synthesize focal stacks is crucial for overcoming the limitations of datasets that lack real focal stacks. To enable robust comparisons with SOTA models and to develop a versatile model for various applications (e.g., AR), we adopt a method to artificially recreate focal stacks from a single image with ground truth depth, similar to \cite{si2023fully}.
The process to artificially recreate focal stacks from a single image (with ground truth depth information) follows these steps: (1) \textbf{Build an arbitrary camera system:} Configure a virtual camera with adjustable focus settings to mimic a physical camera system. (2) \textbf{Define focus distances:} Set specific focus distances to simulate camera focusing at different depths, similar to real-world camera behavior. (3) \textbf{Apply circular kernel for blurring:} Iterate over the image with a circular kernel to add blur based on the ground truth (GT) depth and the defined focus distances.

For the blurring process, we use Equation~\eqref{eq:Coc}, which is the same equation that has been used in recent works~\cite{Defocus_2020_CVPR, si2023fully} for creating the synthesized defocus blur. This equation is used to determine the extent of blur for pixels outside the specific focal plane according to the GT depth.
\begin{equation}
\label{eq:Coc}
    c = \frac{|S_2 - S_1|}{S_2} \frac{f^2}{ N \times (S_1 - f)}, 
\end{equation}

where \( f \) is the camera's focal length, \( N \) is the f-number (aperture) of the lens, \( S_1 \) is the distance to the in-focus subject, and \( S_2 \) is the distance beyond which subjects are considered out of focus. This equation allows for realistic synthesis of focal stacks by adjusting the blur based on depth, transforming a single image into multiple focal stack images. This enables us to leverage single-image depth datasets like \NYU for training.




\subsection{End-To-End Mobile Pipeline}
\label{subsec:e2e_mobile_impl}

We designed a mobile pipeline for utilizing \pipeline for depth estimation on a mobile client. The process begins with the mobile client capturing a focal stack of images(5 or 10), each representing different focus distances of the same scene. 
We developed an Android app using the Camera2 API to efficiently capture these focal stacks, adjusting the focus plane across five different values in a short time frame (approximately $141 \pm 20$ ms on a Pixel 6 Pro). All images are resized to $480 \times 640$ for uniform processing and reducing computational cost.

To address potential misalignment among focal stack images, we leverage the mobile device's built-in optical image stabilization sensor (OIS) during capture. This ensures the focal stack is properly aligned before being sent to our model pipeline. Once the images are captured, they are transmitted to a server equipped with an NVIDIA RTX 4090 GPU for processing. On the server, we first apply OCR-based image alignment using OpenCV, before feeding the focal stack to \pipeline. The resulting dense depth map, which can be utilized in various mobile applications such as AR, is then returned to the mobile device.

\section{Experiments}

A key challenge in evaluating \pipeline is the lack of commonly used datasets that can allow us to directly compare with single-image and DFF-based methods. To address this, we evaluated \pipeline's performance across four different datasets. 
For direct comparison with other DFF-based models, we utilized the \ddff dataset. Additionally, we synthesized the \NYU dataset as described in Section~\ref{sec:data_synthesize} to facilitate comparisons between single-image methods and DFF-based approaches. To assess the generalizability of \pipeline, we conducted zero-shot evaluations on \ArkitScenes(quantitatively, qualitatively), as well as \mobiledepth (qualitatively). 

\pipeline consistently outperforms existing methods on all datasets. For example, \pipeline achieves a 6.1\% and a 36\% improvement in RMSE on the \ddff and \NYU datasets compared to each dataset's specific SOTA model (Tables \ref{tab:performance_ddff}, \ref{tab:performance_nyu_SIDE}, \ref{tab:performance_nyu_DFF}), respectively. Additionally, \pipeline demonstrates superior generalizability in zero-shot evaluations. For example, \pipeline achieves a 82.7\% improvement in RMSE compared to Depth Anything (Table~\ref{tab:ARkitScene_performance_zero-shot}). Furthermore, our qualitative experiments (Figures~\ref{fig:comparison_ARkitScenes}, and \ref{fig:MobileDepth_vis}) show that \pipeline produces higher-quality depth maps compared to other models. Please refer to Supplementary Material for more visualizations and results for inference time.

\subsection{Datasets}

We select a diverse set of datasets, including real-world and synthetic datasets, to comprehensively evaluate \pipeline and compare against single-image and DFF-based methods. 
The \ddff dataset offers real-world focal stacks captured by a light-field camera, enabling direct comparison with other DFF-based models. 
The \mobiledepth dataset, similar to \ddff, contains real-world DFF data captured using a mobile phone and is used for qualitative comparison. 
We also transform the following two datasets with synthesized focal stacks to enable more comparisons.
\NYU is a widely recognized benchmark for monocular depth estimation, particularly for indoor scenes, which helps us compare \pipeline with other single-image depth estimation models. 
Lastly, \ArkitScenes is a large-scale, diverse dataset captured with mobile devices that represents real-world challenges in depth estimation~\cite{10.1145/3638550.3641122}. Please refer to Supplementary Material for more details about datasets.





\subsection{Evaluation Metrics}
We evaluate the metric depth accuracy with the following metrics: MSE as $\frac{1}{M} \sum_{i=1}^{M} (d_i - \hat{d}_i)^2$, RMSE as $\sqrt{\frac{1}{M} \sum_{i=1}^{M} (d_i - \hat{d}_i)^2}$, and AbsRel Error as $\frac{1}{M} \sum_{i=1}^{M} \left| \frac{d_i - \hat{d}_i}{d_i} \right|$. 
Here, \(d_i\) and \(\hat{d}_i\) denote the ground truth and predicted depth at pixel \(i\), and \(M\) is the total number of pixels in the image.
Additionally, we assess the accuracy at threshold values using \( \delta_1 \), \( \delta_2 \), and \( \delta_3 \) which measure the percentage of pixels where the predicted depth \(\hat{d}_i\) is within \(1.25\), \(1.25^2\), and \(1.25^3\) times the ground truth depth \(d_i\), respectively. 
For the \ddff dataset, we use the same metrics for disparity calculation. 



\subsection{Comparison to the State-of-the-Art}
\begin{figure}[t]
    \centering
        \includegraphics[width=\columnwidth]{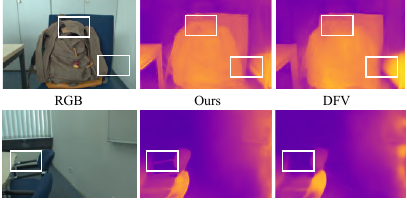}
    \caption{\pipeline performance in capturing small details in depth maps in comparison to DFV on \ddff.}
\label{fig:detail_dfv_comparison}
\vspace{-5pt}
\end{figure}

\begin{table}[t]
\caption{Performance comparison on the \ddff dataset. \textbf{Bold} values represent the best results.
All numbers for other works have been taken from the DFV paper. The unit for all metrics is disparity.
}
\label{tab:performance_ddff}
\resizebox{\columnwidth}{!}{
\begin{tabular}{@{}l|rrrrrrr@{}}
\toprule
\textbf{Model} & \textbf{MSE} $\downarrow$ & \textbf{RMSE} $\downarrow$ & \textbf{AbsRel} $\downarrow$ & \textbf{$\delta_1$} $\uparrow$ & \textbf{$\delta_2$} $\uparrow$ & \textbf{$\delta_3$} $\uparrow$\\ \midrule

RDF~\cite{7271087} & {$91.8 \times 10^{-4}$} & {$0.0941$} & {1.00} & {$0.16$} & {0.33} & {0.47}\\
Defocus-Net~\cite{Defocus_2020_CVPR} & $8.6 \times 10^{-4}$ & $0.0255$ & {0.17} & $0.61$ & 0.94 & {$0.97$}\\
DDFF~\cite{hazirbas18ddff} & $8.9 \times 10^{-4}$ & $0.0276$ & $0.24$ & $0.61$ & $0.88$ & $0.96$\\
DFFintheWild~\cite{won2022learning} & {$5.7 \times 10^{-4}$} & - & {0.17} & {0.78} & $0.87$ & $0.94$\\
DFV~\cite{3eb3b6e31bc743c68f1e7439bd4f4799} & {$5.7 \times 10^{-4}$} & {$0.0213$} & {0.17} & {$0.76$} & {0.94} & {0.98}\\

\midrule
\textbf{Ours} & $\bm{5.1 \times 10^{\bm{-4}}}$ & \textbf{0.0200} & \textbf{0.17} & \textbf{0.79} & \textbf{0.95} & \textbf{0.98}\\
\bottomrule
\end{tabular}
}
\end{table}

\begin{table}[t]
\centering
\caption{Performance comparison on the \NYU dataset with other Depth from Focus/Defocus methods. 
\textbf{Bold} values represent the best results. 
The evaluation uses an upper bound of 10 meters on the ground truth depth map.
DefocuseNet~\cite{Defocus_2020_CVPR} results have been taken from the corresponding paper. The unit for all metrics is disparity.
}
\label{tab:performance_nyu_DFF}
\resizebox{\columnwidth}{!}{
\begin{threeparttable}
\begin{tabular}{@{}l|lrrrrrrr@{}}
\toprule
\textbf{Model} & \textbf{Type\tnote{$\star$}} & \textbf{RMSE} $\downarrow$ & \textbf{AbsRel} $\downarrow$ & \textbf{$\delta_1$} $\uparrow$ & \textbf{$\delta_2$} $\uparrow$ & \textbf{$\delta_3$} $\uparrow$\\ 
\midrule
DefocusNet\cite{Defocus_2020_CVPR} & DFD & 0.493 & - & - & - & - \\
DefocusNet ($\leq2m$)\tnote{$\ddagger$} & DFD & 0.180 & - & - & - & - \\
DFV~\cite{3eb3b6e31bc743c68f1e7439bd4f4799} & DFF & 0.136 & 0.028 & \textbf{0.996} & 1.000 & 1.000\\
\midrule
\textbf{Ours} & DFF & \textbf{0.128} & \textbf{0.026} & \underline{}{0.995} & \textbf{1.000} & \textbf{1.000} \\
\textbf{Ours} ($\leq 2m$)\tnote{$\ddagger$} & DFF & {0.082} & 0.034 & 0.988 & 0.998 & 1.000 \\

\bottomrule
\end{tabular}
\begin{tablenotes}
    \item[$\ddagger$]  These rows show performance metrics for distances under 2 meters.
    \item[$\star$] \emph{DFD/DFF} stand for depth from defocus/focus depth estimation.
  \end{tablenotes}
\end{threeparttable}
    }

\vspace{-10pt}
\end{table}

\begin{table}[t]
\centering
\caption{Performance comparison on the \NYU dataset with single-image depth estimation models. 
\textbf{Bold} and \underline{underlined} values represent the best and second-best results. 
The evaluation uses an upper bound of 10 meters on the ground truth depth map.
All the numbers for other works have been taken from the corresponding papers.  The unit for all metrics is meter.
}
\label{tab:performance_nyu_SIDE}
\resizebox{\columnwidth}{!}{
\begin{threeparttable}
\begin{tabular}{@{}l|lrrrrrrr@{}}
\toprule
\textbf{Model} & \textbf{Type\tnote{$\star$}} & \textbf{RMSE} $\downarrow$ & \textbf{AbsRel} $\downarrow$ & \textbf{$\delta_1$} $\uparrow$ & \textbf{$\delta_2$} $\uparrow$ & \textbf{$\delta_3$} $\uparrow$\\ 
\midrule
ZoeDepth~\cite{https://doi.org/10.48550/arxiv.2302.12288}\tnote{$\dagger$} & SIDE & 0.270 & 0.075 & 0.96 & 0.995 & 0.999 \\
VPD~\cite{zhao2023unleashing} & SIDE & 0.254 & 0.069 & 0.96 & 0.995 & 0.999 \\
ECoDepth~\cite{patni2024ecodepth} & SIDE & 0.218 & 0.059 & 0.97 & 0.997 & 0.999 \\
Depth Anything~\cite{depthanything} & SIDE & \underline{0.206} & \underline{0.056} & \underline{0.98} & \underline{0.998} & \textbf{1.000} \\
\midrule
\textbf{Ours} & DFF & \textbf{0.128} & \textbf{0.026} & \textbf{0.99} & \textbf{1.000} & \textbf{1.000} \\

\bottomrule
\end{tabular}
\begin{tablenotes}
    \item[$\star$] \emph{SIDE} stand for single image depth estimation.
    \item[$\dagger$]  For ZoeDepth we have used ZoeDepth-M12-N version.
  \end{tablenotes}
\end{threeparttable}
    }
\end{table}

\para{Results on \ddff.} This dataset presents a significant challenge for depth-from-focus (DFF) methods due to large texture-less areas, where focus cues are often weak in the focal stack, leading to increased error possibilities. As shown in Table~\ref{tab:performance_ddff}, our model outperforms the current state-of-the-art models, achieving a 10.5\% improvement in MSE and a 6.1\% improvement in RMSE compared to DFV~\cite{3eb3b6e31bc743c68f1e7439bd4f4799}. Additionally, compared to Defocus-Net~\cite{Defocus_2020_CVPR}, our model achieves a 40.7\% improvement in MSE and a 21.6\% improvement in RMSE. These improvements highlight the effectiveness of our combination of Depth Anything and DFV with the refinement layer, which helps address scale inaccuracies and handle texture-less regions with weak focus cues. Qualitative results, as shown in Figure~\ref{fig:detail_dfv_comparison}, further demonstrate that our model produces more detailed and higher-quality depth maps, preserving structural information from Depth Anything. This illustrates the strength of our pipeline in leveraging Depth Anything’s depth cues and improving upon DFF models through scale refinement.

\para{Results on \NYU.} 
Table~\ref{tab:performance_nyu_DFF} compares \pipeline with DFF-based methods. We trained the DFV module on the \NYU dataset and used it as a baseline for comparison. With the addition of our refinement layer, \pipeline outperforms DFV by 5.9\% in RMSE and 7.1\% in AbsRel. Furthermore, compared to DefocusNet~\cite{Defocus_2020_CVPR}, \pipeline achieves a 74.0\% improvement in RMSE. These results highlight the effectiveness of fusing Depth Anything and DFV with our novel refinement layer.

Table~\ref{tab:performance_nyu_SIDE} compares \pipeline with single-image depth estimation methods on the \NYU dataset. 
The significant performance gaps above single-image approaches highlight the effectiveness of \pipeline. 
Specifically, compared to Depth Anything~\cite{depthanything}, \pipeline improves RMSE by 37.9\%. It also outperforms diffusion-based models like ECoDepth~\cite{patni2024ecodepth} by 41.3\% in RMSE. These results demonstrate that incorporating focal stack cues into the depth estimation task leads to substantial gains over single-image depth methods. 

\subsection{Zero-Shot Evaluation}
We evaluated \pipeline's zero-shot performance, which is essential for most depth model applications, by comparing it against baselines on the \ArkitScenes and \mobiledepth datasets.
We show that \pipeline outperforms SOTA models on the \ArkitScenes dataset and produces more detailed depth maps while preserving object boundaries on the \mobiledepth. Additionally, \pipeline demonstrates better consistency in depth estimations across two different zoom levels, addressing the common scale ambiguity problem faced by single-image depth estimation models. Please check Supplementary Material for more visual comparisons of \pipeline to ARCore~\cite{ARCore_website} and DFV.

\begin{figure}[t]
    \centering
        \includegraphics[width=\columnwidth]{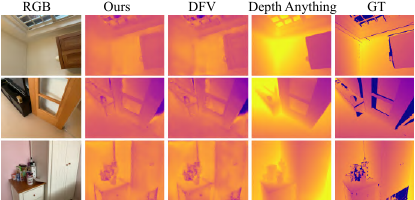}
    \caption{\pipeline's zero-shot performance on \ArkitScenes compared to DFV and Depth Anything, demonstrating improved depth accuracy and detail preservation. }
\label{fig:comparison_ARkitScenes}
\end{figure}

\begin{figure}[t]
    \centering
        \includegraphics[width=\columnwidth]{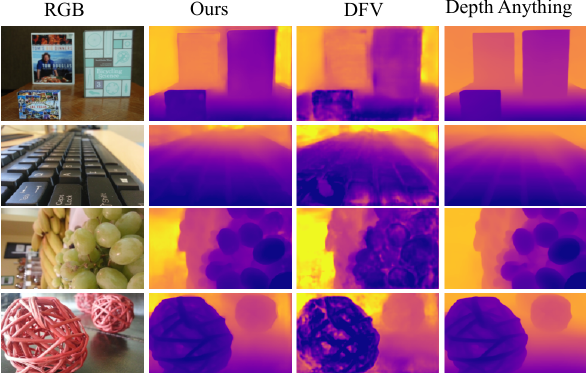}
    \caption{Qualitative results on \mobiledepth dataset.
    }
\label{fig:MobileDepth_vis}
\end{figure}

\para{Results on \ArkitScenes.}
As shown in Table~\ref{tab:ARkitScene_performance_zero-shot}, \pipeline achieves a 32.6\% improvement in RMSE over DFV and a 45.3\% improvement over Depth Anything, while using a smaller model size. When trained on \NYU dataset, \pipeline achieves the best RMSE results and comparable AbsRel performance, highlighting the effectiveness of focal stack cues in mobile AR scenarios. 

Additionally, as seen in the comparison between the two versions of our model, one trained on \NYU and the other on \ddff, the model trained on \NYU demonstrates better zero-shot performance, improving RMSE by 19.4\% compared to the one trained on \ddff. 
This suggests that training with more data, even synthetic ones like \NYU, can improve zero-shot performance in real-world scenarios. Figure~\ref{fig:comparison_ARkitScenes} further illustrates \pipeline's performance in capturing details and producing high-quality depth maps. Also, note that Depth Anything overestimates the depth values for some regions.

\para{Results on \mobiledepth.}
Figure~\ref{fig:MobileDepth_vis} shows sample qualitative comparisons on \mobiledepth (other samples are shown in Supp.). 
We can see that \pipeline maintains higher levels of detail and object boundary accuracy. 
For example, in the last row, \pipeline captures the details on the ball correctly, but DFV's output is noisy. Comparing our results with Depth Anything, we can see that \pipeline successfully utilizes all the details captured by Depth Anything. More examples are available in Supplementary Material.

\begin{table}[t]

\centering
\caption{Zero-shot evaluation comparison on the \ArkitScenes validation set with a focal stack size of 5. \textbf{Bold} represents the best results. \underline{Underline} represents second best results. The unit for all metrics is meter.
}
\label{tab:ARkitScene_performance_zero-shot}
\resizebox{0.95\columnwidth}{!}{
\begin{threeparttable}
\begin{tabular}{@{}l|lccr}
\toprule
\textbf{Model} & \textbf{Type\tnote{$\star$}} & \textbf{RMSE} $\downarrow$ & \textbf{AbsRel} $\downarrow$  & \textbf{\#Params}\\
\midrule
ZoeDepth~\cite{https://doi.org/10.48550/arxiv.2302.12288}\tnote{$\dagger$} & SIDE & 0.61 & \underline{0.33} &334.82M\\
DistDepth~\cite{wu2022toward} & SIDE & 0.94 & 0.45 & 68M\\
ZeroDepth~\cite{tri-zerodepth}  & SIDE & 0.62  & {0.37} & 233M \\
Depth Anything~\cite{depthanything}   & SIDE & 0.53  & \textbf{0.32} & 335.79M \\
DFV~\cite{3eb3b6e31bc743c68f1e7439bd4f4799}  & DFF & 0.43  & 0.51 & 15M \\
\midrule
\textbf{Ours (\NYU)} & DFF & \textbf{0.29} & 0.42 & {65.6M} \\
\textbf{Ours (\ddff)} & DFF & \underline{0.36} & 0.49 & {65.6M} \\
\bottomrule
\end{tabular}

\begin{tablenotes}
    \item[$\star$] \emph{DFD} stands for depth from defocus/focus depth estimation. \emph{SIDE} stands for single image depth estimation.
    \item[$\dagger$]  For ZoeDepth we have used ZoeDepth-M12-N version.
\end{tablenotes}

\end{threeparttable}
}
\vspace{-5pt}
\end{table}

\begin{figure*}[t]
    \centering
    \begin{subfigure}[b]{0.48\textwidth}
        \centering
        \includegraphics[width=\textwidth]{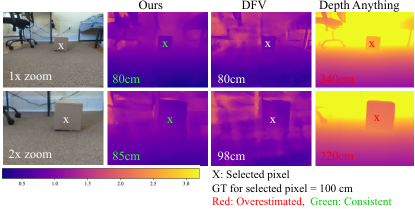}
        \caption{}
        \label{fig:scaleAmbiguity_a}
    \end{subfigure}
    \hfill
    \begin{subfigure}[b]{0.48\textwidth}
        \centering
        \includegraphics[width=\textwidth]{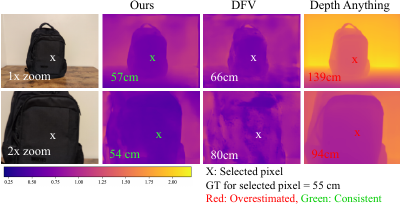}
        \caption{}
        \label{fig:scaleAmbiguity_b}
        
    \end{subfigure}
    \caption{
    Comparison of depth estimations across different zoom levels for various models. 
    \pipeline achieves more accurate and consistent estimations than DFV and Depth Anything. Specifically, DFV's estimations are more noisy and have larger discrepancies (e.g., 18 cm) between two zoom levels than \pipeline (e.g., 5 cm).  
    Depth Anything vastly overestimates the depth of both scenes. We captured the images using our own focal stack capturing mobile app (\S\ref{subsec:e2e_mobile_impl}) using a Google Pixel 6 Pro and obtained the GT depth using a measuring tape.
    }
    \label{fig:scaleAmbiguity}
\end{figure*}

\para{Effect of different zoom levels} \label{subsec
} We evaluated \pipeline, DFV, and Depth Anything across two different zoom levels using two scenes (Figure~\ref{fig:scaleAmbiguity}). \pipeline maintains consistent depth estimations in both scenes. In the first scene (Figure~\ref{fig:scaleAmbiguity_a}), \pipeline shows only a 5 cm difference between two zoom levels, while DFV exhibits a larger discrepancy of 18 cm and produces noisier results. Depth Anything significantly overestimates depth, with errors up to 240 cm, and shows inconsistencies of up to 120 cm between two zoom levels. In the second scene (Figure~\ref{fig:scaleAmbiguity_b}), \pipeline achieves an impressive depth difference of just 3 cm, while Depth Anything has a 45 cm difference and DFV shows a 14 cm difference, both failing to provide a correct and consistent depth.

\subsection{Ablation Study}
We investigate the impact of several key design choices. See more ablation studies in Supplementary Material.

\para{Effect of Uncertainty Map.}
Table~\ref{tab:confidence_performance} shows the performance comparison of \pipeline with and without the uncertainty map on the \ddff dataset. When the uncertainty map is included, model achieves an 8.6\% reduction in MSE and a 2.4\% reduction in RMSE. These improvements suggest that incorporating the uncertainty map helps the refinement layer to refine depth estimates more accurately, leading to more accurate estimations.

\begin{table}[t]
\caption{Performance comparison between \pipeline trained with and without uncertainty map for refinement layer on \ddff. The unit for all metrics is disparity.}
\label{tab:confidence_performance}
\resizebox{\columnwidth}{!}{
\begin{tabular}{@{}c|rrrrrr@{}}
\toprule
     \textbf{Uncertainty Map} & \textbf{MSE} $\downarrow$ & \textbf{RMSE} $\downarrow$ & \textbf{AbsRel} $\downarrow$ & \textbf{$\delta_1$} $\uparrow$ & \textbf{$\delta_2$} $\uparrow$ & \textbf{$\delta_3$} $\uparrow$\\ \midrule
     \xmark & ${5.6 \times 10^{{-4}}}$ & {0.0205} & {0.168} & {0.78} & {0.94} & {0.98}\\
     \midrule
     \cmark & ${5.1 \times 10^{{-4}}}$ & {0.0200} & {0.170} & {0.79} & {0.95} & {0.98}\\
\bottomrule
\end{tabular}
}
\end{table}

\para{Effect of Refinement Layer.}
Table~\ref{tab:ablation_refinement} shows the performance of \pipeline on the \ddff and \NYU datasets with and without the refinement layer, using a focal stack size of 5. The addition of the refinement layer improves depth estimation performance on both datasets. For \ddff, the RMSE improves by 10.7\% and AbsRel by 10.5\%. For the \NYU dataset, the refinement layer leads to a significant RMSE reduction of 76.8\%, with AbsRel improving by 85.6\%, demonstrating a substantial boost in accuracy.

\begin{table}[t]
\caption{Effect of refinement layer on \NYU and \ddff with focal stack size of 5. The unit for \ddff is disparity and for \NYU is meter.}
\resizebox{\columnwidth}{!}{
\begin{tabular}{@{}l|crrr@{}}
\toprule
\textbf{Method} & \textbf{Dataset} &\textbf{RMSE} $\downarrow$ & \textbf{AbsRel} $\downarrow$ & \textbf{$\delta_1$} $\uparrow$  \\ \midrule
Globally Scaled & \multirow{2}{*}{\ddff} & {0.0224} & 0.19 & 0.72 \\
Globally Scaled +  Refinement&  & 0.0200& 0.17 & 0.79 \\
\midrule
Globally Scaled & \multirow{2}{*}{\NYU} & {0.552} & 0.18 & 0.77\\
Globally Scaled + Refinement & & 0.128& 0.03 & 0.99 \\
\bottomrule
\end{tabular}
}
\label{tab:ablation_refinement}
\vspace{-10pt}
\end{table}

\section{Conclusion and Future Work}
Achieving robust and accurate metric depth estimation in the wild remains a challenging and important problem. 
Recent work demonstrated that single-image depth models such as ZoeDepth have poor generalizability in challenging real-world environments~\cite{10.1145/3638550.3641122}. Concurrently, we observe that existing foundational models like Depth Anything still face significant issues with scale ambiguity. 
Focal stack, data that has become increasingly available on mobile devices, has the potential to provide valuable depth information to address the scale ambiguity problem. 
These observations motivate our design of \pipeline, a novel end-to-end metric depth estimation pipeline. 

At the core, \pipeline synergistically fuses metric information from the focal stack and depth prior from a foundational model, and uses a refinement model to further enhance details. 
\pipeline establishes new SOTA results on the commonly used DFF dataset \ddff, improving RMSE and MSE over DFV~\cite{3eb3b6e31bc743c68f1e7439bd4f4799} by 6.1\% and 10.5\%, respectively. 
On another real focal stack dataset, \mobiledepth, \pipeline achieves strong zero-shot performance. Similarly, on datasets with synthetic focal stack (\NYU and \ArkitScenes), \pipeline outperforms both DFF and single-image SOTA models.
This robust performance is achieved with only mobile cameras, making \pipeline highly practical and accessible compared to solutions that rely on specialized hardware like LiDAR or ToF sensors. 

As part of future work, we will explore improving both the DFF modules and the single-image depth prior. 
While we showcase stronger generalization capabilities compared to previous methods, there is still a substantial performance drop on out-of-domain samples. We would like to close the gap by scaling up training and better utilizing synthetic data.

\section*{Acknowledgement}{
This work was supported in part by NSF Grants \#2350189, \#2346133, and \#2236987.
}

\balance
{\small
\bibliographystyle{ieee_fullname}
\bibliography{ref}
}

\clearpage
\appendix

\section{Implementation}
\label{sec:implementation}

\subsection{Training Details}
\label{sec:supplementary_training}

For the \NYU dataset, we set weight decay (\(\lambda\)) to 0.001 and use a learning rate of \(3 \times 10^{-4}\). The batch size is 24, and we use the original data size (\(480 \times 640\)) without any resizing.

For \ddff, weight decay (\(\lambda\)) is set to 0.0001, with a learning rate of \(1 \times 10^{-4}\). The batch size is 8, and the input size during training is \(224 \times 224\) pixels, with random crop and flip augmentations applied. For evaluation, the original image size of \(383 \times 552\) is used, following DFF-based methods~\cite{3eb3b6e31bc743c68f1e7439bd4f4799}. Focal stacks are arranged in ascending order of focal distance to ensure consistency in depth processing.

For the refinement layer, we initialize the MiDaS-small encoder backbone with pre-trained ImageNet~\cite{5206848} weights, while the remaining layers are randomly initialized to allow adaptation to our depth estimation task.

\section{Experiments}
\label{sec:extra_experiments}

\subsection{Dataset}
\label{subsec:dataset_apendix}
\para{\ddff}\cite{hazirbas18ddff}. We follow the dataset split specified in DFV~\cite{3eb3b6e31bc743c68f1e7439bd4f4799}. 
The training set consists of six scenes, each containing 100 samples, while the test set includes six different scenes with 20 samples per scene. Each sample contains a 10-frame focal stack along with a corresponding ground truth disparity map. The images have a resolution of 383 × 552 pixels. For our training and evaluation, we used a focal stack of 5 frames, similar to DFV~\cite{3eb3b6e31bc743c68f1e7439bd4f4799}.

\para{\mobiledepth}~\cite{Suwajanakorn_2015_CVPR} includes 11 aligned focal stacks from 11 different scenes. The image resolutions range from 360 × 640 to 518 × 774, with each stack containing between 14 and 33 frames. 
Since ground truth depth and focal distance are not provided, we used this dataset solely for qualitative comparisons on aligned focal stack images.

\para{\NYU}~\cite{Silberman:ECCV12} contains over 24K densely labeled RGB and depth image pairs in the training set and 654 pairs in the test set. This dataset covers a broad range of indoor environments, with ground truth depth maps obtained using a structured light sensor, provided at a resolution of 640 × 480 pixels.

\para{\ArkitScenes}~\cite{dehghan2021arkitscenes} is a large-scale dataset designed for mobile AR applications.
For our experiments,  we utilized a subset of 5.6K images for evaluating \pipeline's zero-shot performance.
This subset provides a comprehensive basis for evaluating the robustness and accuracy of our model under real-world AR conditions.

\subsection{Model Performance Analysis}
\label{subsec:latency_analysis}
We conducted a performance analysis to demonstrate the efficiency of our model compared to SOTA models like ZoeDepth-M12-N, Depth Anything, and DFV. All tests were performed on an Nvidia RTX 4090 GPU. Table~\ref{tab:Latency_analysis} shows that \pipeline achieves an inference time of 20 ms, which is 4.3X faster than ZoeDepth-M12-N and 2.85X faster than Depth Anything. 
Additionally, our model’s size is 5.3X smaller than ZoeDepth-M12-N and 5.2X smaller than Depth Anything. Despite being more compact, \pipeline provides a considerable improvement in performance and is highly suitable for deployment on devices with limited memory and storage.  While DFV is faster at 8 ms and smaller in size, previous sections have shown that its depth estimation accuracy is significantly lower.

\begin{table}[h]
\centering
\caption{Performance analysis of the three SOTA models on Nvidia RTX 4090 with \ddff. Note: We use the ViT Large version for Depth Anything.}
\label{tab:Latency_analysis}
\resizebox{\columnwidth}{!}{
\begin{tabular}{@{}l|ccr@{}}
\toprule
\textbf{Model} & \textbf{Inference Time} & \textbf{Size}   & \textbf{\#Params}\\
\midrule
ZoeDepth-M12-N~\cite{https://doi.org/10.48550/arxiv.2302.12288} & $86 \pm 6$ ms & 1.28 GB & 344.82M \\
Depth Anything~\cite{depthanything} & $57 \pm 5$ ms & 1.25 GB & 335.79M\\
{DFV} & {8 ± 2 ms} & {0.07 GB} & {15M} \\
{\textbf{Ours}} & {20 ± 2 ms} & {0.24 GB} & {65.6M} \\
\bottomrule
\end{tabular}
}
\end{table}

\subsection{Qualitative Comparison}
\begin{figure}[h]
    \centering
        \includegraphics[width=\columnwidth]{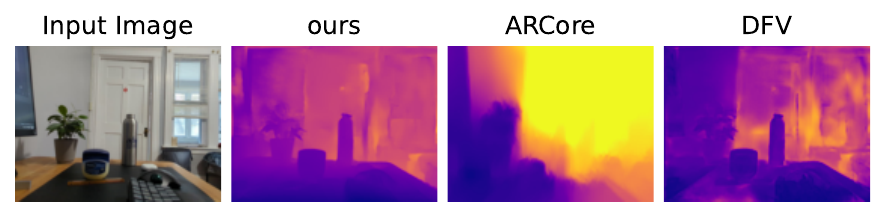}
    \caption{
   Qualitative comparison with ARCore and DFV. Our model outputs better depth by preserving object boundaries and overall geometrical information about the scene. In our experiments with ARCore, depth maps were obtained by moving the camera around the scene until no further improvement was observed.
    }
\label{fig:realworld_arcore}
\end{figure}

\para{Qualitative Comparison with ARCore and DFV.}
Depth estimation plays a crucial role in augmented reality (AR) applications, where accurate depth maps are essential for tasks such as rendering occlusions and precise object placement. We compared our model against the depth maps generated by the commercial ARCore framework~\cite{ARCore_website} and DFV~\cite{3eb3b6e31bc743c68f1e7439bd4f4799}. Utilizing an Android app, we captured a focal stack of five images and sent it over WiFi to an edge server for alignment and inference. Figure~\ref{fig:realworld_arcore} shows that our model preserves better edge details and object boundaries compared to ARCore, while also producing smoother and more reliable depth maps than DFV.

\begin{figure*}[h]
    \centering
        \includegraphics[width=\textwidth]{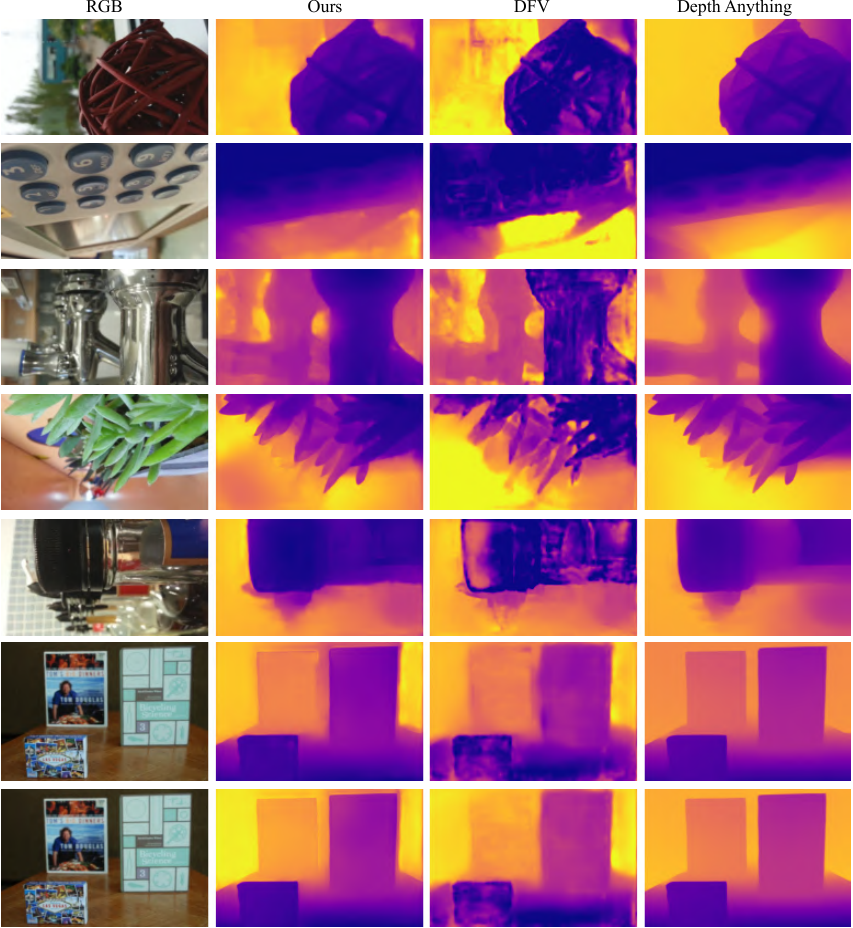}
    \caption{Additional qualitative results on the \mobiledepth dataset. The focal stacks in rows 6,7 are taken from the same scenes with different camera motions, therefore have slightly different frame alignment
    }
\label{fig:MobileDepth_vis_extra}
\end{figure*}

\para{Qualitative Comparison on \mobiledepth.} Figure~\ref{fig:MobileDepth_vis_extra} presents additional results on the aligned scenes of the \mobiledepth dataset. All deep learning methods generalize well to these scenes without fine-tuning. In row 4, our method successfully captures intricate details in the plants, and in the last row, \pipeline provides smoother and more accurate depth estimations, even capturing the depth behind objects. However, our model struggles with depth estimation for transparent surfaces, such as glass. The focal stacks in rows 6,7 are taken from the same scenes with different camera motions, therefore have slightly different frame alignment. We refer readers to ~\cite{Suwajanakorn_2015_CVPR} for more details of this dataset. Overall, \pipeline consistently delivers smoother depth maps with better boundary preservation compared to other methods.

\begin{figure*}[h]
    \centering
        \includegraphics[width=\textwidth,keepaspectratio]{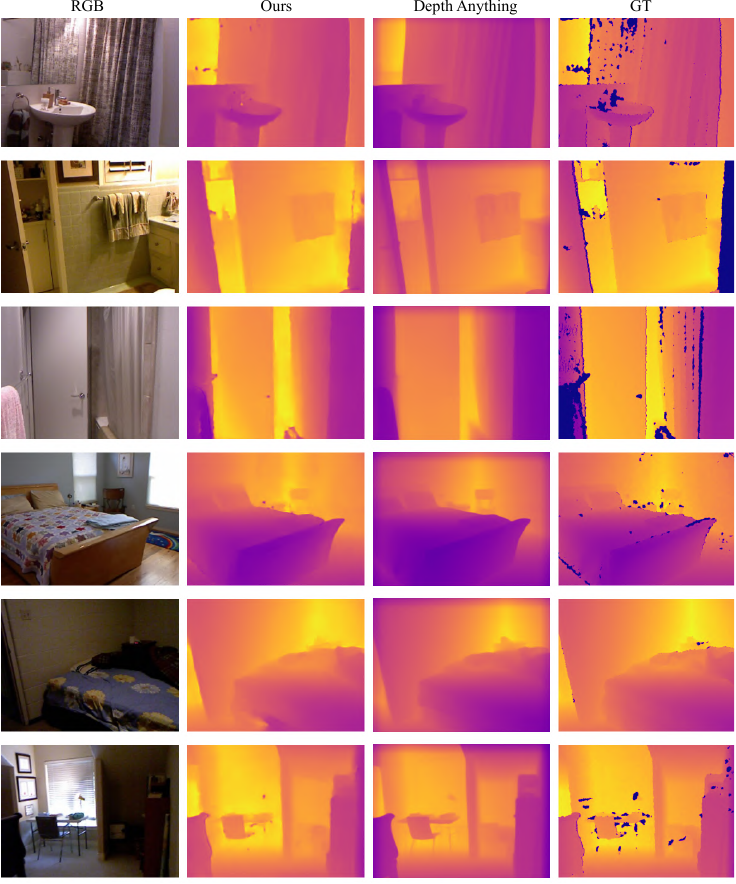}
    \caption{Qualitative results on the \NYU dataset.}
\label{fig:nyu_vis}
\end{figure*}

\para{Qualitative Comparison on \NYU.}
Figure~\ref{fig:nyu_vis} compares our model with Depth Anything on the \NYU dataset. Both models generate accurate depth maps; however, our model excels at capturing depth for distant objects more closely aligned with the ground truth, as seen in rows 3 and 6. Additionally, our model captures finer details more effectively, particularly in row 2.

\subsection{Ablation Study}
\label{sec: Ablation_extra}
\para{Effect of Focal Stack Size.}
We analyzed the effect of focal stack size on \pipeline's performance across the \NYU, \ddff, and \ArkitScenes datasets (Table \ref{tab:focalstackSize}). On the \NYU dataset, increasing the focal stack size from 5 to 10 reduced the RMSE by 35.2\% and the AbsRel by 42.3\%, while both configurations still achieved state-of-the-art (SOTA) results. Similarly, on the \ArkitScenes dataset, using a focal stack size of 10 slightly reduced the RMSE by 10.3\%, confirming that \pipeline's performance benefits from a larger focal stack size but remains robust even with smaller stacks. The performance difference on the \ddff dataset was negligible between stack sizes, demonstrating consistent accuracy across different configurations.

\begin{table}[h]

\centering
\caption{Effect of focal stack size on \pipeline. Both focal stack sizes yield new SOTA results, and there are no significant performance differences between these two settings.}
\label{tab:focalstackSize}
\resizebox{\columnwidth}{!}{
\begin{tabular}{@{}c|ccrr}
\toprule
    \textbf{Focal Stack Size} &\textbf{Trained} &\textbf{Evaluated} &\textbf{RMSE} $\downarrow$ & \textbf{AbsRel} $\downarrow$  \\
\midrule
5 & \NYU & \NYU & 0.128 & 0.026\\ 
10 & \NYU & \NYU& 0.083 & 0.015 \\
\midrule
5 & \ddff & \ddff & 0.0200 &0.1695 \\ 
10 & \ddff & \ddff& 0.0200 &0.1690 \\
\midrule
5  & \NYU &\ArkitScenes & 0.29 & 0.42\\ 
10 & \NYU &\ArkitScenes & 0.29 & 0.39 \\
\bottomrule
\end{tabular}
}
\end{table}

\para{Different Global Scaling Methods.}
We evaluated the performance of various global scaling (GS) methods on the \ddff dataset, as shown in Table~\ref{tab:abblation_gs}. The least square method showed competitive performance, achieving results comparable to more complex method RANSAC, but with a significant computational advantage. For example, it was over 30x faster than RANSAC with 200 iterations and 50 sample size, while providing similar accuracy with only a 1.8\% increase in RMSE compared to the best RANSAC configuration. This makes the least square method the most efficient choice for global scaling, ensuring reliable depth estimates without adding considerable overhead.

\begin{table}[h]
\caption{Comparison of global scaling (GS) methods on the \ddff dataset.}
\resizebox{\columnwidth}{!}{
\begin{tabular}{@{}l|rrrr@{}}
\toprule
\textbf{Method} & \textbf{RMSE} $\downarrow$ & \textbf{AbsRel} $\downarrow$ & \textbf{$\delta_1$} $\uparrow$ & \textbf{Time} (ms) $\downarrow$ \\ \midrule
Least Square & {0.0224} & 0.19 & 0.72 & {3} \\
RANSAC (itr: 60, Sample size: 5) & 0.0246 & 0.19 & 0.73 & 34 \\
RANSAC (itr: 100, Sample size: 20) & 0.0236 & 0.18 & {0.76} & 96 \\
RANSAC (itr: 200, Sample size: 50) & {0.0228} & {0.17} & 0.75 & 170 \\
\bottomrule
\end{tabular}
}
\label{tab:abblation_gs}

\end{table}

\end{document}